\documentclass[10pt,twocolumn,letterpaper]{article}
\usepackage{cvpr}
\usepackage{times}
\usepackage{epsfig}
\usepackage{graphicx}
\usepackage{amsmath}
\usepackage{amssymb}
\usepackage{adjustbox}

\usepackage{times}
\usepackage{helvet}
\usepackage{courier}
\usepackage{algorithm}
\usepackage{algorithmic}
\usepackage{color}
\usepackage{graphicx} 
\usepackage{subfigure}
\usepackage{graphicx}
\usepackage{epsfig}
\usepackage{epsf}

\usepackage[pagebackref=true,breaklinks=true,letterpaper=true,colorlinks,bookmarks=false]{hyperref}

\cvprfinalcopy 

\ifcvprfinal\fi
\begin{document}

\title{Learning Deep Structure-Preserving Image-Text Embeddings}

\author{Liwei Wang\footnotemark[1]\\
{\tt\small lwang97@illinois.edu} 
\and
Yin Li\footnotemark[2]\\
{\tt\small yli440@gatech.edu}
\and 
Svetlana Lazebnik\footnotemark[1]\\
{\tt\small slazebni@illinois.edu}\\
\hspace{-4in} \footnotemark[1] University of Illinois at Urbana-Champaign
\quad \quad \quad \footnotemark[2]  Georgia Institute of Technology
}

\maketitle

\begin{abstract}
This paper proposes a method for learning joint embeddings of images and text using a two-branch neural network with multiple layers of linear projections followed by nonlinearities. The network is trained using a large-margin objective that combines cross-view ranking constraints with within-view neighborhood structure preservation constraints inspired by metric learning literature. Extensive experiments show that our approach gains significant improvements in accuracy for image-to-text and text-to-image retrieval. Our method achieves new state-of-the-art results on the Flickr30K and MSCOCO image-sentence datasets and shows promise on the new task of phrase localization on the Flickr30K Entities dataset.
\end{abstract}

\section{Introduction}

Computer vision is moving from predicting discrete, categorical labels to generating rich descriptions of visual data, for example, in the form of natural language. There is a surge of interest in image-text tasks such as image captioning~\cite{fang2014captions,karpathy2014deepcvpr,karpathy2014deep,kiros2014unifying,mao2014deep,socher2014grounded,vinyals2014show,xu2015show} and visual question answering~\cite{antol2015vqa,gao2015you,yu2015visual}. A core problem for these applications is how to measure the semantic similarity between visual data (e.g., an input image or region) and text data (a sentence or phrase). A common solution is to learn a joint embedding for images and text into a shared latent space where vectors from the two different modalities can be compared directly. This space is usually of low dimension and is very convenient for cross-view tasks such as image-to-text and text-to-image retrieval. 

Several recent embedding methods~\cite{gong2014multi,gong2014improving,klein2014fisher} are based on Canonical Correlation Analysis (CCA)~\cite{hardoon2004canonical}, which finds linear projections that maximize the correlation between projected vectors from the two views. Kernel CCA~\cite{hardoon2004canonical} is an extension of CCA in which maximally correlated nonlinear projections, restricted to reproducing kernel Hilbert spaces with corresponding kernels, are found. Extensions of CCA to a deep learning framework have also been proposed~\cite{andrew2013deep,mikolajczyk2015deep}. However, as pointed out in~\cite{ma2015finding}, CCA is hard to scale to large amounts of data. In particular, stochastic gradient descent (SGD) techniques cannot guarantee a good solution to the original generalized eigenvalue problem, since covariance estimated in each small batch (due to the GPU memory limit) is extremely unstable.

An alternative to CCA is to learn a joint embedding space using SGD with a ranking loss. WSABIE~\cite{weston2011wsabie} and DeVISE \cite{frome2013devise} learn linear transformations of visual and textual features to the shared space using a {\em single-directional} ranking loss that applies a margin-based penalty to incorrect annotations that get ranked higher than correct ones for each training image. Compared to CCA-based methods, this ranking loss easily scales to large amounts of data with stochastic optimization in training. 
As a more powerful objective function, a few other works have proposed a {\em bi-directional} ranking loss that, in addition to ensuring that correct sentences for each training image get ranked above incorrect ones, also ensures that for each sentence, the image described by that sentence gets ranked above images described by other sentences~\cite{karpathy2014deepcvpr,karpathy2014deep,kiros2014unifying,socher2014grounded}. However, to date, it has proven frustratingly difficult to beat CCA with an SGD-trained embedding: Klein et al.~\cite{klein2014fisher} have shown that properly normalized CCA~\cite{gong2014multi} on top of state-of-the-art image and text features can outperform considerably more complex models.

Another strand of research on multi-modal embeddings is based on deep learning~\cite{ba2015predicting,kiros2014multimodal,kiros2014unifying,mao2014deep,ngiam2011multimodal,srivastava2012multimodal}, utilizing such techniques as deep Boltzmann machines~\cite{srivastava2012multimodal}, autoencoders~\cite{ngiam2011multimodal}, LSTMs~\cite{donahue2014long}, and recurrent neural networks~\cite{mao2014deep,venugopalan2014translating}. By making it possible learn nonlinear mappings, deep methods can in principle provide greater representational power than methods based on linear projections~\cite{frome2013devise,gong2014improving,klein2014fisher,weston2011wsabie}. 

\begin{figure}
\hspace*{-2.4cm}  
\includegraphics[width=1.7\linewidth]
{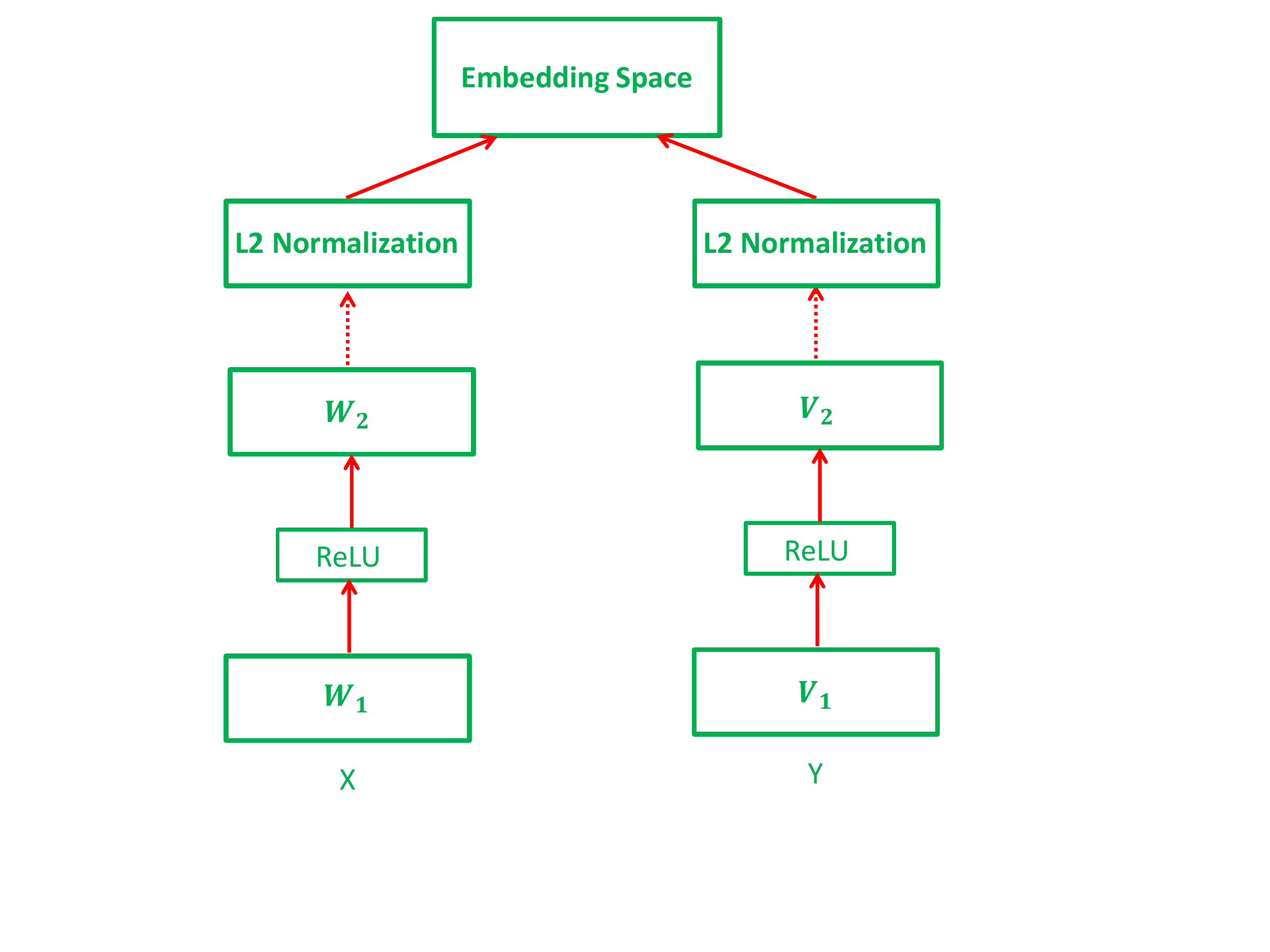}
\vspace{-1.6cm}
\caption{Our model structure: there are two branches in the network, one for images ($X$) and the other for text ($Y$). Each branch consists of fully connected layers with ReLU nonlinearities between them, followed by L2 normalization at the end. }
\label{fig:Model_structure}
\end{figure}

In this work, we propose to learn an image-text embedding using a two-view neural network with two layers of nonlinearities on top of any representations of the image and text views (Figure \ref{fig:Model_structure}). These representations can be given by the outputs of two pre-trained networks, off-the-shelf feature extractors, or trained jointly end-to-end with the embedding. To train this network, we use a bi-directional loss function similar to~\cite{karpathy2014deepcvpr,karpathy2014deep,kiros2014unifying,socher2014grounded}, combined with constraints that preserve neighborhood structure within each individual view. Specifically, in the learned latent space, we want images (resp. sentences) with similar meaning to be close to each other. Such within-view structure preservation constraints have been extensively explored in the metric learning literature~\cite{hu2014discriminative,mensink2012metric,shaw2011learning,shaw2009structure,weinberger2005distance,vzbontar2014computing}. In particular, the Large Margin Nearest Neighbor (LMNN) approach~\cite{weinberger2005distance} tries to ensure that for each image its target neighbors from the same class are closer than samples from other classes. As our work will show, these constraints can also provide a useful regularization term for the cross-view matching task.

From the viewpoint of architecture, our method is similar to the two-branch Deep CCA models~\cite{andrew2013deep,mikolajczyk2015deep}, though it avoids Deep CCA's training-time difficulties associated with covariance matrix estimation. Our network also gains in accuracy by performing feature normalization (L2 and batch normalization) before the embedding loss layer. Finally, our work is related to deep similarity learning~\cite{bell2015learning,bromley1993signature,chopra2005learning,han2015matchnet,hoffer2014deep,schroff2015facenet,wang2014learning}, though we are solving a cross-view, not a within-view, matching problem. Siamese networks for similarity learning (e.g., \cite{schroff2015facenet}) can be considered as special cases of our framework where the two views come from the same modality and the two branches share weights.

Our proposed approach substantially improves the state of the art for image-to-sentence and sentence-to-image retrieval on the Flickr30K~\cite{young2014image} and MSCOCO~\cite{lin2014microsoft} datasets. We are also able to obtain convincing improvements over CCA on phrase localization for the Flickr30K Entities dataset~\cite{plummer2015flickr30k}. 
 
\section{Deep Structure-Preserving Embedding}

Let $X$ and $Y$ denote the collections of training images and sentences, each encoded according to their own feature vector representation.
We want to map the image and sentence vectors (which may have different dimensions initially) to a joint space of common dimension. We use the inner product over the embedding space to measure similarity, which is equivalent to the Euclidean distance since the outputs of the two embeddings are L2-normalized. In the following, $d(x,y)$ will denote the Euclidean distance between image and sentence vectors in the embedded space.

\subsection{Network Structure}

We propose to learn a nonlinear embedding in a deep neural network framework. As shown in Figure \ref{fig:Model_structure}, our deep model has two branches, each composed of fully connected layers with weight matrices $W_l$ and $V_l$. Successive layers are separated by Rectified Linear Unit (ReLU) nonlinearities. We apply batch normalization~\cite{ioffe2015batch} right after the last linear layer. And at the end of each branch, we add L2 normalization.

In general, each branch can have a different number of layers, and if the inputs of the two branches $X$ and $Y$ are produced by their own networks, the parameters of those networks can be trained (or fine-tuned) together with the parameters of the embedding layers. However, in this paper, we have obtained very satisfactory results by using two embedding layers per branch on top of pre-computed image and text features (see Section \ref{sec:features} for details). 

\subsection{Training Objective} \label{sec:loss}

Our training objective is a stochastic margin-based loss that includes bidirectional cross-view ranking constraints, together with within-view structure-preserving constraints. 
\smallskip

\noindent \textbf{Bi-directional ranking constraints.} Given a training image $x_i$,
let $Y^{+}_{i}$ and $Y^{-}_{i}$ denote its sets of matching (positive) and non-matching (negative) sentences, respectively. We want the distance between $x_i$ and each positive sentence $y_j$ to be smaller than the distance between $x_i$ and each negative sentence $y_k$ by some enforced margin $m$:
\begin{equation} 
d(x_i, y_j) + m < d(x_i, y_k) \quad \forall y_j \in Y^{+}_i, \forall y_k \in Y^{-}_{i} \,. \label{eq:img2sen}
\end{equation}
Similarly, given a sentence $y_{i'}$, we have
\begin{equation} 
d(x_{j'}, y_{i'}) + m < d(x_{k'}, y_{i'}) \quad \forall x_{j'} \in X^{+}_{i'}, \forall x_{k'} \in X^{-}_{i'},\label{eq:sen2img}
\end{equation}
where $X^{+}_{i'}$ and $X^{-}_{i'}$ denote the sets of matching (positive) and non-matching (negative) images for $y_{i'}$. 
\smallskip

\noindent \textbf{Structure-preserving constraints.} Let $N(x_{i})$ denote the neighborhood of $x_i$ containing images that share the same meaning. In our case, this is the set of images described by the same sentence as $x_i$. Then we want to enforce a margin of $m$ between $N(x_i)$ and any point outside of the neighborhood:
\begin{equation}
d(x_i, x_j) + m < d(x_i, x_k) \quad \forall x_j \in N(x_{i}), \forall x_k \not\in N(x_{i}),\label{eq:structure_x}
\end{equation}

Analogously to (\ref{eq:structure_x}), we define the constraints for the sentence side as
\begin{equation}
d(y_{i'}, y_{j'}) + m < d(y_{i'}, y_{k'}) \quad \forall y_{j'} \in N(y_{i'}), \forall y_{k'} \not\in N(y_{i'}),\label{eq:structure_y}
\end{equation}
where $N(y_{i'})$ contains sentences describing the same image. 

\begin{figure}
\hspace*{-.5cm}
\includegraphics[width=1.1\linewidth]{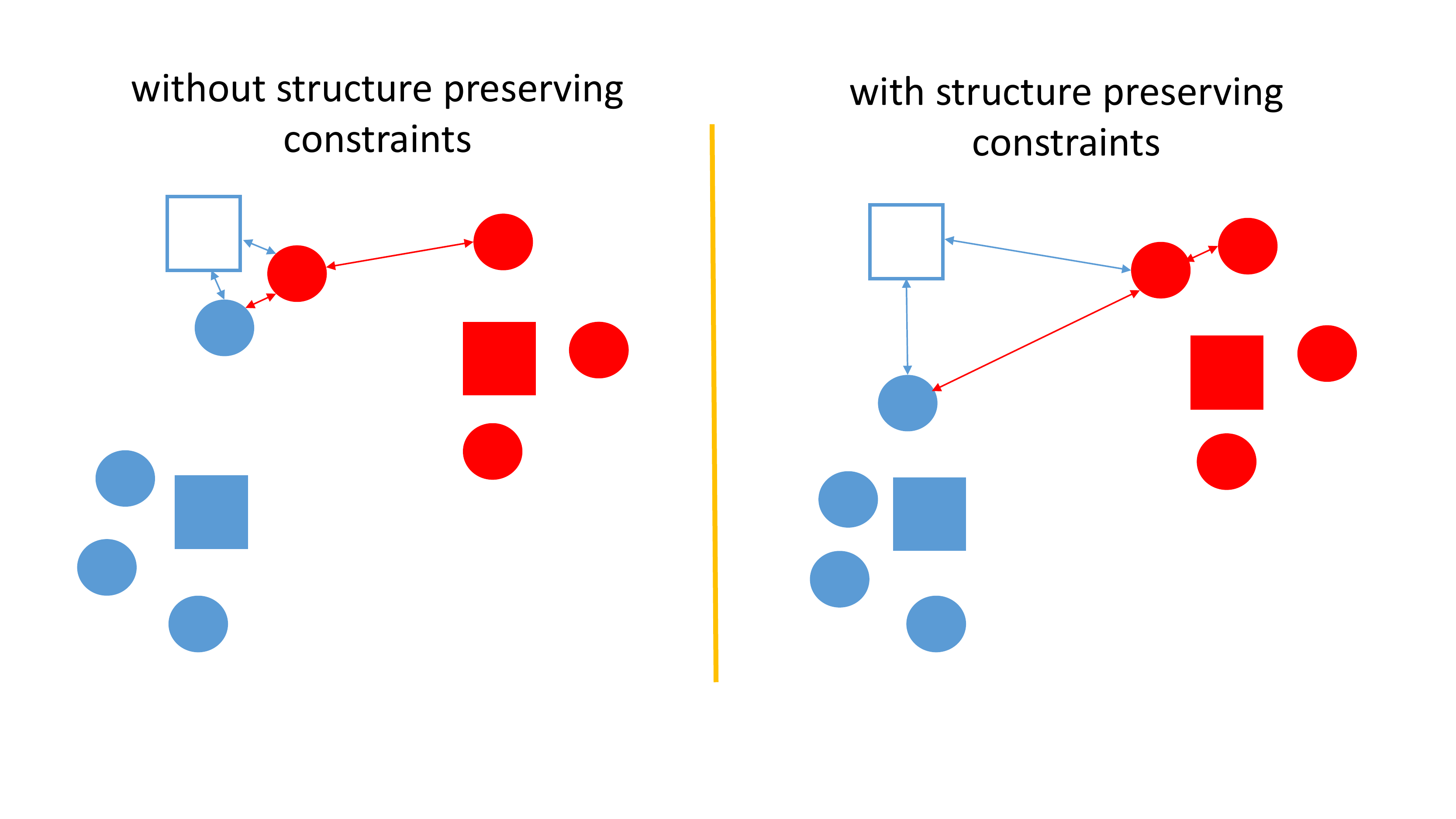}
\vspace*{-1cm}
\caption{Illustration of the proposed structure-preserving constraints for joint embedding learning (see text). Rectangles represent images and circles represent sentences. Same color indicates matching images and sentences.}
\label{fig:preserveStruct}
\end{figure}

Figure \ref{fig:preserveStruct} gives an intuitive illustration of how within-view structure preservation can help with cross-view matching. The embedding space on the left satisfies the cross-view matching property. That is, each square (representing an image) is closer to all circles of the same color (representing its corresponding sentences) than to any circles of the other color. Similarly, for any circle (sentence), the closest square (image) has the same color. However, for the new image query (white square), the embedding space gives an ambiguous matching result since both red and blue circles are very close to it. This problem is mitigated in the embedding on the right, where within-view structure constraints are added, pushing semantically similar sentences (same color circles) closer to each other. 

Note that our two image-sentence datasets, Flickr30K and MSCOCO, consist of images paired with five sentences each. The neighborhood of each image, $N(x_i)$, generally only contains $x_i$ itself, since it is rare for two different images to be described by an identical sentence. Thus, the image-view constraints (eq. \ref{eq:structure_x}) are trivial, while the neighborhood of each sentence $N(y_{i'})$ has five members. However, for the region-phrase dataset of Section \ref{sec_rp}, many phrases have multiple region exemplars, so we get a non-trivial set of constraints for the image view. \smallskip

\noindent \textbf{Embedding Loss Function.} We convert the constraints to our training objective in the standard way using hinge loss. The resulting loss function is given by 
\begin{equation}\label{eq:obj}
\begin{split}
L(X, Y) = & \sum_{i,j,k} \max[0, m + d(x_i, y_j) - d(x_i, y_k)]  \\
+ & \lambda_1 \sum_{i',j',k'} \max[0, m + d(x_{j'}, y_{i'}) - d(x_{k'}, y_{i'})] \\
+ & \lambda_2 \sum_{i,j,k} \max[0, m + d(x_i, x_j) - d(x_i, x_k) ]  \\
+ & \lambda_3 \sum_{i',j',k'} \max[0, m + d(y_{i'}, y_{j'}) - d(y_{i'}, y_{k'}) ] \,,   \\
\end{split} 
\end{equation}
where the sums are over all triplets defined as in the constraints (1-4).
The margin $m$ could be different for different types of distance or even different instances. But to make it easy to optimize, we fix $m$ for all terms across all training samples ($m = 0.1$ in the experiments). The weight
$\lambda_1$ balances the strengths of both ranking terms. In other work with a bi-directional ranking loss~\cite{karpathy2014deepcvpr,karpathy2014deep,kiros2014unifying,socher2014grounded}, this is always set to 1, but in our case, we found $\lambda_1=2$ produces the best results. The weights $\lambda_2,\lambda_3$ control the importance of the structure-preserving terms, which act as regularizers for the bi-directional retrieval tasks. We usually set both to small values like 0.1 or 0.2 (see Section \ref{sec:experiments} for details).
\smallskip

\noindent \textbf{Triplet sampling}. Our loss involves all triplets consisting of a target instance, a positive match, and a negative match. Optimizing over all such triplets is computationally infeasible. Therefore, we sample triplets within each mini-batch and optimize our loss function using SGD. Inspired by~\cite{joachims2009cutting,shaw2011learning}, instead of choosing the most violating negative match in all instance space, we select top $K$ most violated matches in each mini-batch. This is done by computing pairwise similarities between all $(x_i, y_j)$, $(x_i, x_j)$ and $(y_i, y_j)$ within the mini-batch. For each positive pair (i.e., a ground truth image-sentence pair, two neighboring images, or two neighboring sentences), we then find at most top $K$ violations of each relevant constraint (we use $K=50$ in the implementation, although most pairs have many fewer violations). Theoretical guarantees of such a sampling strategy have been discussed in~\cite{shaw2011learning}, though not in the context of deep learning. In our experiments, we observe convergence within 30 epochs on average.

In Section \ref{sec:experiments}, we will demonstrate the performance of our method both with and without structure-preserving constraints. For training the network without these constraints, we randomly sample 1500 pairs ($x_i,y_i$) to form our mini-batches. For the experiments with the structure-preserving constraints, in order to get a non-empty set of constraint triplets, we need a moderate number of positive pairs (i.e., at least two sentences that are matched to the same image) in each mini-batch. However, random sampling of pairs cannot guarantee this. Therefore, for each $x_i$ in a given mini-batch, we add one more positive sentence distinct from the ones that may already be included among the sampled pairs, resulting in mini-batches of variable size. 

\section{Experiments} \label{sec:experiments}

In this section, we analyze the contributions of different components of our method and evaluate it on image-to-sentence and sentence-to-image retrieval on popular Flickr30K~\cite{young2014image} and MSCOCO~\cite{lin2014microsoft} datasets, and on phrase localization on the new Flickr30K Entities dataset~\cite{plummer2015flickr30k}. 

\subsection{Features and Network Settings} \label{sec:features}

In image-sentence retrieval experiments, to represent images, we follow the implementation details in \cite{klein2014fisher,plummer2015flickr30k}. Given an image, we extract the 4096-dimensional activations from the 19-layer VGG model~\cite{simonyan2014very}. Following standard procedure, the original $256\times256$ image is cropped in ten different ways into $224\times224$ images: the four corners, the center, and their x-axis mirror image. The mean intensity is then subtracted from each color channel, the resulting images are encoded by the network, and the network outputs are averaged. 

To represent sentences and phrases, we primarily use the Fisher vector (FV) representation~\cite{perronnin2010improving} as suggested by Klein et al.~\cite{klein2014fisher}. Starting with 300-dimensional word2vec vectors~\cite{mikolov2013distributed} of the sentence words, we apply ICA as in~\cite{klein2014fisher} and construct a codebook with 30 centers using both first- and second-order information, resulting in sentence features of dimension $300*30*2 = 18000$. We only use the Hybrid Gaussian-Laplacian mixture model (HGLMM) from \cite{klein2014fisher} for our experiments rather than the combined HGLMM+GMM model which obtained the best performance in \cite{klein2014fisher}. To save memory and training time, we perform PCA on these 18000-dimensional vectors to reduce them to 6000 dimensions. PCA also makes the original features less sparse, which is good for the numerical stability of our training procedure.

Since FV is already a powerful hand-crafted nonlinear transformation of the original sentences, we are also interested in exploring the effectiveness of our approach on top of simpler text representations. To this end, we include results on 300-dimensional means of word2vec vectors of words in each sentence/phrase, and on tf-idf-weighted bag-of-words vectors. For tf-idf, we pre-process all the sentences with WordNet's lemmatizer~\cite{bird2006nltk} and remove stop words. For the Flickr30K dataset, our dictionary size (and descriptor dimensionality) is 3000, and for MSCOCO, it is 5600.

For our experiments using tf-idf or FV text features, we set the embedding dimension to be 512. On the image ($X$) side, when using 4096-dimensional visual features,  $W_1$ is a $4096 \times 2048$ matrix, and $W_2$ is a $2048 \times 512$ matrix. That is, the output dimensions of the two layers are [2048, 512]. On the text ($Y$) side, the output dimensions of the $V_1$ and $V_2$ layers are [2048, 512]. For the experiments using 300-D word2vec features, we use a lower dimension (256) for the embedding space and the intermediate layers output are accordingly changed to [1024, 256].

We train our networks using SGD with momentum $0.9$ and weight decay $0.0005$. We use a small learning rate starting with $0.1$ and decay the learning rate by $0.1$ after every $10$ epochs. To accelerate the training and also make gradient updates more stable, we apply batch normalization~\cite{ioffe2015batch} right after the last linear layer of both network branches. We also use a Dropout layer after ReLU with probability = 0.5. We set the mini-batch size to 1500 ground truth image-sentence pairs and augment these pairs as necessary as described in the previous section. Compared with CCA-based methods, our method has much smaller memory requirements and is scalable to larger amounts of data. 

\begin{table*}[t]
\centering
\small
\begin{center}
\begin{tabular}{|l|l|l|l|l|l|l|l|}
\hline
& Methods on Flickr30K                         & \multicolumn{3}{l|}{Image-to-sentence}                                                                                     & \multicolumn{3}{l|}{Sentence-to-image}                                                                                        \\ \hline
&                                             & R@1                                    & R@5                                    & R@10                                   & R@1                                    & R@5                                    & R@10                                   \\ \hline
(a) State of the art 


& Deep CCA~\cite{mikolajczyk2015deep} & 27.9 & 56.9 & 68.2&  26.8  &  52.9 & 66.9  \\ 
& mCNN(ensemble)~\cite{ma2015multimodal} & 33.6  & 64.1  & 74.9 &  26.2  & 56.3   & 69.6 \\ 
& m-RNN-vgg~\cite{mao2014deep} & 35.4   & 63.8   &73.7   & 22.8   & 50.7   & 63.1  \\ 
& Mean vector~\cite{klein2014fisher}   & 24.8 & 52.5 & 64.3 & 20.5 & 46.3 & 59.3 \\ 
& CCA (FV HGLMM)~\cite{klein2014fisher}  &  34.4 & 61.0 & 72.3 & 24.4 & 52.1 & 65.6     \\
& CCA (FV GMM+HGLMM)~\cite{klein2014fisher}  & 35.0   & 62.0  & 73.8  & 25.0 & 52.7  & 66.0   \\ 
& CCA (FV HGLMM)~\cite{plummer2015flickr30k} & 36.5 & 62.2 & 73.3 & 24.7 & 53.4 & 66.8  \\  
\hline

(b) Fisher vector & Linear + one-directional  & 33.5  & 61.7 & 73.6  & 21.0  & 47.4 & 60.5  \\ 
& Linear + bi-directional  & 34.6  & 64.3  & 74.9 & 24.2 & 52.0  & 64.2    \\ 
& Linear + bi-directional + structure  & 35.2  & 66.8  & 76.2   & 25.6   & 54.8  & 66.5 \\ 
& Nonlinear + one-directional  & 37.5  & 65.6  & 76.9  & 22.4  & 50.9 & 63.3  \\ 
& Nonlinear + bi-directional  &  39.3  & 68.0  & 78.3   & 28.1   & 59.2   & 71.2     \\ 
& Nonlinear + bi-directional + structure   &  \textbf{40.3}   &  \textbf{68.9}    & \textbf{79.9}      &  \textbf{29.7}    & \textbf{60.1}   & \textbf{72.1}      \\ 

\hline

(c) Mean vector & Nonlinear + bi-directional  & 33.5 & 60.2  & 71.9  & 22.8  & 52.5  & 65.0    \\ 
& Nonlinear + bi-directional + structure  & 35.7 & 62.9  & 74.4  & 25.1  & 53.9  & 66.5 \\ 

\hline

(d) tf-idf & Nonlinear + bi-directional    & 38.7         & 66.6          &  76.9     & 27.6     &  57.0          &  69.0       \\
& Nonlinear + bi-directional + structure       & 40.1         & 67.6          &  78.2     & 28.1     &  58.5          & 69.8      \\ 

\hline

\end{tabular}
\caption{Bidirectional retrieval results. 
The numbers in (a) come from published papers, and the numbers in (b-d) are results of our approach using different textual features. Note that the Deep CCA results in~\cite{mikolajczyk2015deep} were obtained with AlexNet~\cite{krizhevsky2012imagenet}. The results of our method with AlexNet are still about 3\% higher than those of~\cite{mikolajczyk2015deep} for image-to-sentence retrieval and 1\% higher for sentence-to-image retrieval.}
\label{table:flickr30k}
\end{center}
\end{table*}

\subsection{Image-sentence retrieval}

In this section, we report results on image-to-sentence and sentence-to-image retrieval on the standard Flickr30K~\cite{young2014image} and MSCOCO~\cite{lin2014microsoft} datasets. Flickr30K \cite{young2014image} consists of 31783 images accompanied by five descriptive sentences each. The larger MSCOCO dataset~\cite{lin2014microsoft} consists of 123000 images, also with five sentences each. 

For evaluation, we follow the same protocols as other recent work~\cite{karpathy2014deepcvpr,klein2014fisher,plummer2015flickr30k}. For Flickr30K, given a test set of 1000 images and 5000 corresponding sentences, we use the images to retrieve sentences and vice versa, and report performance as Recall@$K$ ($K=1,5,10$), or the percentage of queries for which at least one correct ground truth match was ranked among the top $K$ matches. For MSCOCO, consistent with~\cite{karpathy2014deepcvpr,klein2014fisher}, we also report results on 1000 test images and their corresponding sentences.

For Flickr30K, bidirectional retrieval results are listed in Table \ref{table:flickr30k}. Part (a) of the table summarizes the performance reported by a number of competing recent methods. In Part (b) we demonstrate the impact of different components of our model by reporting results for the following variants.

\begin{itemize}
  \item Linear + one-directional: In this setting, we keep only the first layers in each branch with parameters $W_1,V_1$, immediately followed by L2 normalization. The output dimensions of $W_1$ and $V_1$ are changed to be the embedding space dimension. In the objective function (eq. \ref{eq:obj}), we set $\lambda_1=0,\lambda_2=0,\lambda_3=0$, only retaining the image-to-sentence ranking constraints. This results in a model similar to WSABIE~\cite{weston2011wsabie}.
  \item Linear + bi-directional: The model structure is as above, and in eq. (\ref{eq:obj}), we set $\lambda_1 = 2,\lambda_2=0,\lambda_3=0$. This form of embedding is similar to~\cite{karpathy2014deepcvpr,karpathy2014deep,kiros2014unifying,socher2014grounded} (though the details of the representations used by those works are quite different).
  \item Linear + bi-directional + structure: same linear model, eq. (\ref{eq:obj}) with $\lambda_1 = 2,\lambda_2=0,\lambda_3=0.2$.
  \item Nonlinear + one-directional: Network as in Figure \ref{fig:Model_structure}, eq. (\ref{eq:obj}) with $\lambda_1=0,\lambda_2=0,\lambda_3=0$.
  \item Nonlinear + bi-directional: Network as in Figure \ref{fig:Model_structure}, eq. (\ref{eq:obj}) with $\lambda_1 = 2,\lambda_2=0,\lambda_3=0$.
  \item Nonlinear + bi-directional + structure: Network as in Figure \ref{fig:Model_structure}, eq. (\ref{eq:obj}) with $\lambda_1 = 2,\lambda_2=0,\lambda_3=0.2$.
\end{itemize}

\begin{table*}[t]
\centering
\small
\begin{center}
\begin{tabular}{|l|l|l|l|l|l|l|l|}
\hline
& Methods on MSCOCO 1000 testing set                         & \multicolumn{3}{l|}{Image-to-sentence}                                                                                     & \multicolumn{3}{l|}{Sentence-to-image}                                                                                        \\ \hline
&                                             & R@1                                    & R@5                                    & R@10                                   & R@1                                    & R@5                                    & R@10                                   \\ \hline
(a) State of the art 

& Mean vector~\cite{klein2014fisher}  & 33.2  & 61.8  & 75.1  & 24.2  &  56.4  & 72.4   \\ 
& CCA (FV HGLMM)~\cite{klein2014fisher} &  37.7 & 66.6  & 79.1  &  24.9  & 58.8  & 76.5   \\ 
& CCA (FV GMM+HGLMM)~\cite{klein2014fisher} & 39.4  & 67.9  & 80.9  & 25.1  & 59.8   & 76.6   \\  
& DVSA~\cite{karpathy2014deepcvpr}& 38.4 & 69.9 & 80.5 & 27.4 & 60.2 & 74.8  \\
& m-RNN-vgg~\cite{mao2014deep} &  41.0    & 73.0     & 83.5    & 29.0    & 42.2    & 77.0   \\
& mCNN(ensemble)~\cite{ma2015multimodal}   &  42.8  & 73.1  & 84.1   & 32.6   & 68.6  & 82.8    \\

\hline

(b) Fisher Vector & Nonlinear+bi-directional  &  47.5     &  77.6   & 88.3      &  36.8    & 72.2  & 85.6   \\
& Nonlinear+bi-directional+structure  &  {\bf 50.1} & {\bf 79.7} & {\bf 89.2}  &  {\bf 39.6} & {\bf 75.2} & {\bf 86.9}      \\ 
\hline

(c) Mean Vector  
& Nonlinear+bi-directional & 39.6       &  74.0         &   84.8     &  32.0    & 67.3         & 81.6    \\ 
& Nonlinear+bi-directional+structure     &  40.7         & 74.2            &  85.3       &  33.5      &  68.7        &  83.2             \\ 
\hline

(d) tf-idf & Nonlinear+bi-directional     &  45.3   & 77.6     & 86.8   &  35.4    & 70.2       & 83.4  \\         
& Nonlinear+bi-directional+structure     &  46.7    & 77.9    &  87.7   & 36.2    & 72.3    & 84.7     \\ 
\hline
\end{tabular}
\caption{Bidirectional retrieval results on MSCOCO 1000-image test set. 
}
\label{table:microsoftCOCO1000_table}
\end{center}
\end{table*}

Note that in all the above configurations we have $\lambda_2=0$, that is, the structure-preserving constraint associated with the image space is inactive, since in the Flickr30K and MSCOCO datasets we do not have direct supervisory information about multiple images that can be described by the same sentence. However, our results for the region-phrase dataset of Section \ref{sec_rp} will incorporate structure-preserving constraints on both spaces.

From Table \ref{table:flickr30k} (b), we can see that changing the embedding function from linear to nonlinear improves the accuracy by about 4\% across the board. Going from one-directional to bi-directional constraints improves the accuracy by 1-2\% for image-to-sentence retrieval and by a bigger amount for sentence-to-image retrieval. Finally, adding the structure-preserving constraints provides an additional improvement of 1-2\% in both linear and nonlinear cases. The methods from Table \ref{table:flickr30k} (a) most comparable to ours are CCA (HGLMM)~\cite{klein2014fisher,plummer2015flickr30k}, since they use the same underlying feature representation with linear CCA. Our linear model with all the constraints of eq. (\ref{eq:obj}) does not outperform linear CCA, but our nonlinear one does.

Finally, to check how much our method relies on the power of the input features, parts (c) and (d) of Table \ref{table:flickr30k} report results for our nonlinear models with and without structure-preserving constraints applied on top of weaker text representations, namely mean of word2vec vectors of the sentence and tf-idf vectors, as described in Section \ref{sec:features}. Once again, we can see that structure-preserving constraints give us an additional improvement. Our results with mean vector are considerably better than the CCA results of~\cite{klein2014fisher} on the same feature, and are in fact comparable with the results of~\cite{klein2014fisher,plummer2015flickr30k} on top of the more powerful FV representation. For tf-idf, we achieve results that are just below our best FV results, showing that we do not require a highly nonlinear feature as an input in order to learn a good embedding. Another possible reason why tf-idf performs so strongly may be that word2vec features are pre-trained on an unrelated text corpus, so they may not be as well adapted to our specific data. 

For MSCOCO, results on 1000 test images are listed in Table \ref{table:microsoftCOCO1000_table}. The trends are the same as in Table \ref{table:flickr30k}: adding structure-preserving constraints on the sentence space consistently improves performance, and our results with the FV text feature considerably exceed the state of the art. We have also tried fine-tuning the VGG network by back-propagating our loss function through all the VGG layers, and obtained about 0.5\% additional improvement.

\begin{table*}
\centering
\label{praselocalization} \setlength{\tabcolsep}{1.5mm} 
\begin{tabular}{|l|l|c|l|c|c|}
\hline
Methods          
     & R@1 & R@5 & R@10
      & mAP(all) 
     \\ \hline
CCA baseline 
&   40.11        &    61.52    &      67.17    &   41.96
\\ \hline
\multicolumn{5}{|l|}{Our method without negative mining}       
     \\ \hline

(a) $\lambda_1 = 2$,~~$\lambda_2 = 0$,~~$\lambda_3 = 0$ 
& 35.83    & 60.51   & 66.70
   &  40.50
\\ 

(b) $\lambda_1 = 2$,~~$\lambda_2 = 0$,~~$\lambda_3 = 0.1$ 
& 36.59  & 60.44   & 66.92    
  & 40.85

\\
(c) $\lambda_1 = 2$,~~$\lambda_2 = 0.1$,~~$\lambda_3 = 0$ 
& \textbf{36.74}  & 60.35   & 66.73
   & \textbf{41.22}

\\
(d) $\lambda_1 = 2$,~~$\lambda_2 = 0.1$,~~$\lambda_3 = 0.1$ 
& 36.72  & \textbf{61.14} & \textbf{67.21}
  & 41.13                
\\ \hline 

\multicolumn{5}{|l|}{Fine-tuned with negative mining}\\
\hline




Fine-tuning (a) for 5 epochs
&  41.77  & 63.01  & 68.27   & 46.55 \\

Fine-tuning (b) for 5 epochs
& 43.77 & 64.22 & \textbf{68.84}    
  & 47.38    \\
 
Fine-tuning (c) for 5 epochs
& 42.88 & 63.41 &68.47     
 & 46.78 \\

Fine-tuning (d) for 5 epochs
& \textbf{43.89}  & \textbf{64.46} & 68.66    
 & \textbf{47.72}  

\\ \hline

\end{tabular}
\caption{Phrase localization results on Flickr30K Entities using Fast-RCNN features.  We use 100 EdgeBox proposals, for which the recall upper bound is $R@100 = 76.91$. } \label{phraselocal} 
\end{table*}

\subsection{Phrase Localization on Flickr30K Entities} 
\label{sec_rp}

The recently published Flickr30K Entities dataset~\cite{plummer2015flickr30k} allows us to learn correspondences between phrases and image regions. Specifically, the annotations in this dataset provide links from 244K mentions of distinct entities in sentences to 276K ground truth bounding boxes (some entities consist of multiple instances, such as ``group of people''). We are interested in this dataset because unlike the global image-sentence datasets, it provides many-to-many correspondences, i.e., each region may be described by multiple phrases and each phrase may have multiple region exemplars across multiple images. This allows us to take advantage of structure-preserving constraints on both the visual and textual spaces. 

As formulated in~\cite{plummer2015flickr30k}, the goal of phrase localization is to predict a bounding box in an image for each entity mention (noun phrase) from a caption that goes with that image. For a particular phrase, we perform the search by extracting 100 EdgeBox~\cite{zitnick2014edge} region proposals and scoring them using our embedding.  To get good performance, the best-scoring box should have high overlap with the ground truth region. This can be considered as a ranking problem, and both CCA and our methods can be trained to match phrases and regions. On the other hand, we should realize that this problem is more like detection, where the algorithm should be able to distinguish foreground objects from boxes that contain only background or poorly localized objects. CCA and Deep CCA are not well suited to this scenario, since there is no way to add negative boxes into their learning stage. However, our margin-based loss function makes it possible.

Plummer et al.~\cite{plummer2015flickr30k} reported baseline results for a region-phrase embedding using CCA on top of ImageNet-trained VGG features. Following Rohrbach et al.~\cite{rohrbach2016grounding}, who obtained big improvements on phrase localization using detection-based VGG features, we also use Fast R-CNN features~\cite{girshick2015fast} fine-tuned on a union of the PASCAL 2007 and 2012 train-val sets \cite{everingham2011pascal}. Consistent with~\cite{plummer2015flickr30k}, we do not average multiple crops for region features. For text, in this section we use only the FV feature. Thus, the input dimension of $X$ is 4096 and the input dimension of $Y$ is 6000 as before (reduced by PCA from the original 18000-D FV). We use the two-layer network structure with $[8192,4096]$ as the intermediate layer dimensions on both the $X$ and $Y$ sides (note that on the $X$ side, the intermediate layer actually doubles the feature dimension). 

For our first experiment, we train our embedding without negative mining, using the same positive region-phrase pairs as CCA. For this, we use the same training set as~\cite{plummer2015flickr30k}, which is resampled with at most ten regions per phrase, for a total of 137133 region-phrase pairs, 70759 of which are unique. As in the previous section, we use initial mini-batch size of 1500. But now, for the full version of our objective (eq. \ref{eq:obj}), we augment the mini-batches by sampling not only additional positive phrases for regions, but also additional positive regions for phrases, to make sure that we have as many triplets as possible for structure-preserving constraints on the region side (eq. \ref{eq:structure_x}) and the phrase side (eq. \ref{eq:structure_y}). 

The results of training our model without negative mining for 28 epochs are shown in the top part of Table \ref{phraselocal}. We use the evaluation protocol proposed by~\cite{plummer2015flickr30k}. First, we treat phrase localization as the problem of retrieving instances of a query phrase from a set of region proposals extracted from test images, and report Recall@$K$, or the percentage of queries for which a correct match has rank of at most $K$ (a region proposal is considered to be a correct match if it has IOU of at least 0.5 with the ground-truth bounding box for that phrase). Second, we report average precision (AP) of ranking bounding boxes for each phrase in the test images that contain that phrase, following nonmaximum suppression. The last column of Table \ref{phraselocal} shows mAP over all unique phrases in the test set, with each unique phrase being treated as its own class label. 

Table \ref{phraselocal} (a-d) shows the performance of our bi-directional ranking objective with different combinations of structure terms. We can see that including the structure terms generally gives better results than excluding them, though the effects of turning on each term separately do not differ too much. In large part, this is because of the limited number of structure-preserving constraint triples for each view. In the Flickr30K Entities training set, for all 130K pairs, there are around 70K unique phrases and 80K regions described by a single phrase. This means, that, for most phrases/regions, there are no more than two corresponding regions/phrases. The top line of Table \ref{phraselocal} gives baseline CCA results. For the pre-trained model without using negative mining, our deep embedding has comparable results with CCA on Recall@5 and Recall@10, but lower results on Recall@1. As mentioned earlier, in our past experience we have found CCA to be surprisingly hard to beat with more complex methods~\cite{gong2014improving,plummer2015flickr30k}. 

In order to further improve the accuracy of our embedding, we need to refine it using negative data from background and poorly localized regions. To do this, we take the embedding trained without negative mining, and for each unique phrase in the training set, calculate the distance between this phrase and the ground truth boxes as well as all our proposal boxes. Then we record those ``hard negative'' boxes that are closer to the phrase than the ground truth boxes. For efficiency, we only sample at most 50 hard negative regions for each unique phrase. Next, we continue training our region-phrase model on a training set augmented with these hard negative boxes, using only the bi-directional ranking constraints (eqs. \ref{eq:img2sen} and \ref{eq:sen2img}). We exclude the structure-preserving constraints because they would now be even more severely outnumbered by the bi-directional ranking constraints.

\begin{figure*}
\hspace*{0cm}  
\includegraphics[width=1.0\linewidth]{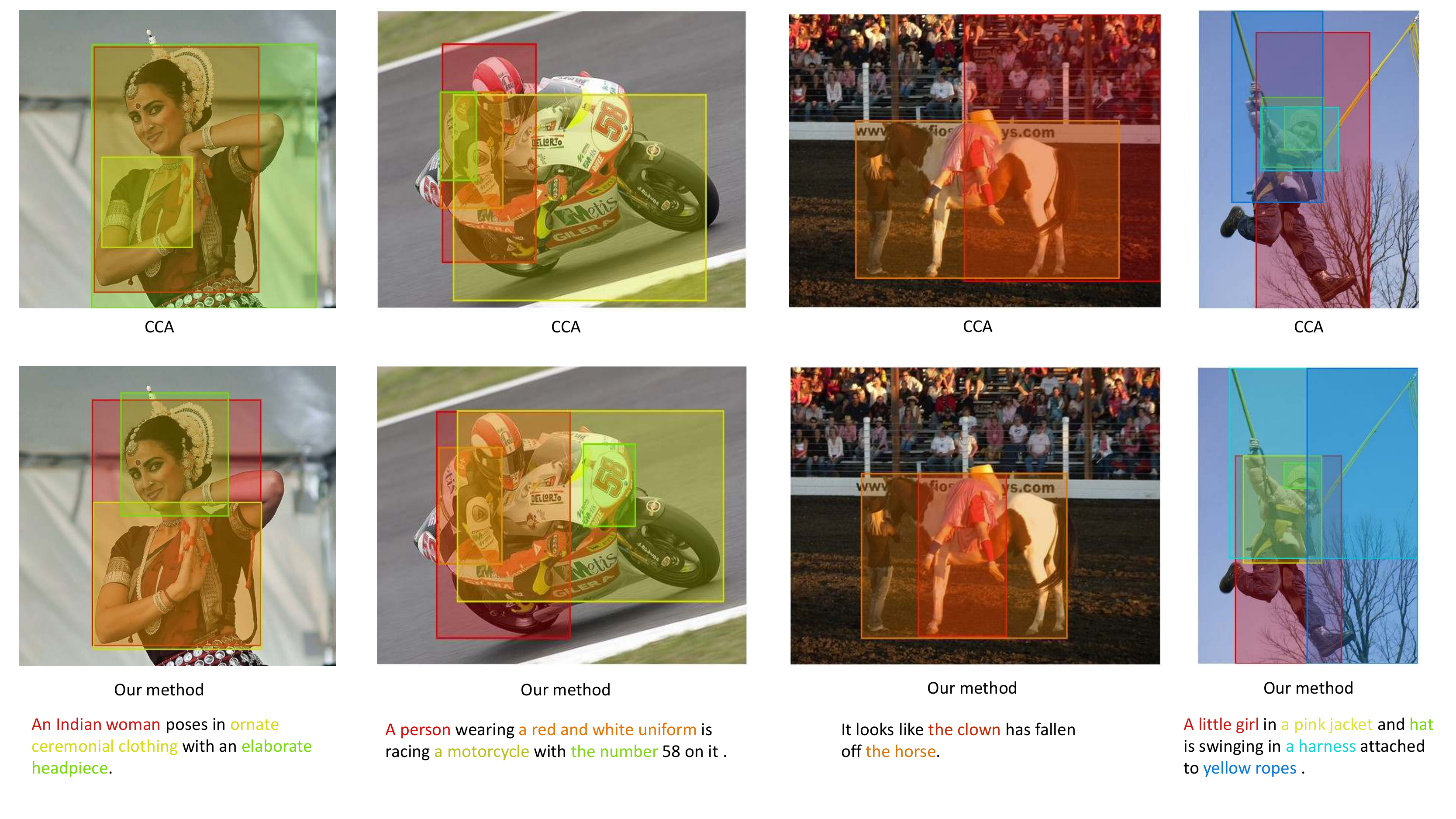}
\vspace{-0.5cm}
\caption{Example phrase localization results. For each image and reference sentence, phrases and best-scoring corresponding regions are shown in the same color. The first row shows the output of the CCA method~\cite{plummer2015flickr30k} and the second row shows the output of our best model (fine-tuned model (d) in Table \ref{phraselocal} with negative mining). For the first (left) example, our method gives more accurate bounding boxes for the clothing and headpiece. For the second example, our method finds the correct bounding box for the number $58$ while CCA completely misses it; for the third column, our method gives much tighter boxes for the horse and clown; and for the last example, our method accurately locates the hat and jacket.}
\label{fig:exampleShow}
\end{figure*}

The last four lines of Table \ref{phraselocal} show the results of fine-tuning the models from Table \ref{phraselocal} (a-d) with hard negative samples. Compared to the best model trained with only positive regions, our Recall@1 and mAP have improved by almost 6\%, and are now considerably better than CCA. Note that in absolute terms, Rohrbach et al.~\cite{rohrbach2016grounding} get higher results, with a R@1 of over 47\%, but they use a much more complex method that includes LSTMs with a phrase reconstruction objective.

Finally, Figure \ref{fig:exampleShow} shows examples of phrase localization in four images where our model improves upon the CCA baseline.

\section{Conclusion}

This paper has proposed an image-text embedding method in which a two-branch network with multiple layers is trained using a margin-based objective function consisting of bi-directional ranking terms and structure-preserving terms inspired by metric learning. Our architecture is simple and flexible, and can be applied to various kinds of visual and textual features. Extensive experiments demonstrate that the components of our system are well chosen and all the terms in our objective function are justified. To the best of our knowledge, our retrieval results on Flickr30K and MSCOCO datasets considerably exceed the state of the art, and we also demonstrate convincing improvements over CCA on the new problem of phrase localization on the Flickr30K Entities dataset.

\section*{Acknowledgments}

This material is based upon work supported by the National Science Foundation under Grant CIF-1302438, Xerox UAC, and the Sloan Foundation. We would like to thank Bryan Plummer for help with phrase localization evaluation.

{\small
\bibliographystyle{ieee}
\bibliography{reference}
}

\end{document}